\documentclass[preprint,11pt, 1p]{elsarticle}

\usepackage{lineno,hyperref}
\usepackage{subfigure}
\usepackage{amsthm,amsmath,amssymb}
\usepackage{mathrsfs}
\usepackage{bm}
\usepackage{algorithm}
\usepackage{algorithmic}
\usepackage{float}
\usepackage{setspace}
\usepackage{supertabular}
\usepackage{longtable}
\usepackage{array}
\usepackage{caption2}
\usepackage{multirow}
\usepackage{makecell}
\usepackage{booktabs}
\usepackage{setspace}

\begin{document}

\begin{frontmatter}



\title{Solving the Data Sparsity Problem in Predicting the Success of the Startups with Machine Learning Methods}


\author[1]{Dafei Yin \corref{cor1}}
\ead{yindafei@chuangxin.com}
%\address[1]{Sinovation Ventures, Dinghao Tower Block B, No. 3 Haidian Street, Haidian District, Beijing, 100080, China}

\author[1]{Jing Li}
\ead{lijing@chuangxin.com}

\author[1]{Gaosheng Wu}
\ead{wugaosheng@chuangxin.com}

\cortext[cor1]{Corresponding author}
\affiliation[1]{organization={Sinovation Ventures},%Department and Organization
            addressline={Dinghao Tower Block B, No. 3 Haidian Street, Haidian District}, 
            %city={},
            postcode={10080}, 
            state={Beijing},
            country={China}}

\begin{abstract}
Predicting the success of startup companies is of great importance for both startup companies and investors. It is difficult due to the lack of available data and appropriate general methods. With data platforms like Crunchbase aggregating the information of startup companies, it is possible to predict with machine learning algorithms. Existing research suffers from the data sparsity problem as most early-stage startup companies do not have much data available to the public. We try to leverage the recent algorithms to solve this problem. We investigate several machine learning algorithms with a large dataset from Crunchbase. The results suggest that LightGBM and XGBoost perform best and achieve 53.03\% and 52.96\% F1 scores. We interpret the predictions from the perspective of feature contribution. We construct portfolios based on the models and achieve high success rates. These findings have substantial implications on how machine learning methods can help startup companies and investors.

\end{abstract}


\end{frontmatter}

% \linenumbers

\section{Introduction}

Predicting the future success of startup companies is of great importance for both startup companies and venture capital (VC) firms.
For startup companies, predicting the future development of themselves and their competitors can help them adjust their development strategies and capture opportunities effectively.
For VC firms, predicting the future success of startup companies helps them balance their profit and risk.

For late-stage companies, the evaluation of future success is mostly based on financial and operating information.
However, for early-stage companies, there is usually not enough data publicly available for prediction.
Traditionally, the evaluation of startup companies
relies heavily on investor's personal experience.
In recent years, machine learning is developing rapidly and achieve great success in many areas.
There exists some research applying machine learning algorithm to predict the future success of startup companies \citep{yankov2014models,mckenzie2017man,arroyo2019assessment,kaiser2020value}, but their methods are not well suited for dealing with sparse data, 
which is common in datasets of startup companies.
With the development of machine learning methods,
recent algorithms like XGBoost \citep{chen2016xgboost} and LightGBM  \citep{ke2017lightgbm} have the potential to solve this data sparsity problem.

In this paper, we aim to make the following three contributions.
First,
we try to leverage the recent progress of machine learning to handle this data sparsity problem.
We validate that the recently developed algorithms, such as XGBoost \citep{chen2016xgboost}, LightGBM  \citep{ke2017lightgbm}, and soft Decision Tree \citep{frosst2017distilling},
outperform many traditional algorithms including Logistic Regression, K Nearest Neighbor, Decision Tree, Multilayer Perceptron, and Random Forests.
Second, 
we construct 19 factors using the data from Crunchbase
\footnote{http://www.crunchbase.com},
 a  data aggregation platform built to track startups on a global scale.
We define multiple time windows to  enrich the number of data samples and take factors like macroeconomy into consideration.
This time window definition is more practical in VC practices.
Third, we introduce interpretability into machine learning models.
We interpret the predictions from the perspective of the contribution of each factor,
finding that company age and past funding experience are the most important factors.

The rest of the paper is structured as follows:
The previous research and theoretical background are reviewed in Sec \ref{sec:literature_review}.
Our approach is introduced in Sec \ref{sec:our_approach}.
Experimental results are discussed in Sec \ref{sec:results}.
The construction of the portfolio is described in Sec \ref{sec:portfolio}.
Sec \ref{sec:conclusion} summarizes our main conclusion
and discusses future research directions.

\section{Theoretical background}
\label{sec:literature_review}

The problem of predicting the future success of startup companies has existed for a long time \citep{schendel1979new,chandler1993measuring}
and is still exploring by scholars \citep{arroyo2019assessment, kaiser2020value}.

Common solutions are classification models based on decisive factors.
Most earlier studies for the prediction of the success of startup companies are based on regression analysis
such as logistic regression \citep{lussier1995nonfinancial,kaiser2020value}, ordered probit model  \citep{lussier2001crossnational} \citep{halabi2014model} \citep{lussier2010three}, and  log-logistic hazard models \citep{holmes2010analysis}.
Researchers also develop expert systems \citep{ragothaman2003predicting} and rule-based methods \citep{yankov2014models}.

In recent years, with the emergence of platforms aggregating business information about companies and the development of machine learning approaches,
it is possible to use machine learning methods to solve the problem of startup prediction.
\citep{yankov2014models} test several tree-based, rule-based, and Bayes-based machine learning methods
based on questionnaires gathered from 142 startup companies in Bulgarian.
The authors suggest that the best-derived model is the tree-based C4.5 \citep{quinlan1993c4}.
\citep{mckenzie2017man} compare the performance of human experts and several machine learning methods,
including Least Absolute Shrinkage and Selection Operator, Support Vector Machines, and Boosted Regression.
They analyze 2,506 firms in a business plan competition in Nigeria.
The author suggests that investors using the combination of man and machine rather than relying on human judges or machine learning-chosen portfolios.
\citep{arroyo2019assessment} analyze the performance of several machine learning methods in a dataset of over 120,000 startup companies retrieved from Crunchbase.
They consider five machine learning algorithms: Support Vector Machines, Decision Tree, Random Forests, Extremely Randomized Trees, and Gradient Tree Boosting.
The results suggest that the Gradient Tree Boosting performs better in predicting the next funding round, while Random Forests and Extremely Randomized Trees perform better in predicting acquisition.
One common problem of the datasets with startup companies is their sparsity nature.
For most early-stage companies, there is usually not very much data available to the public.
Current approaches are not well suited for this problem.

Machine learning is developing rapidly in recent years, and many new models have emerged.
Gradient Boosting Decision Tree (GBDT) \citep{friedman2001greedy} is a highly effective and widely used machine learning method, due to its efficiency, accuracy, and interpretability.
It has several effective implementations recently, including XGBoost \citep{chen2016xgboost} and LightGBM \citep{ke2017lightgbm}.
XGBoost \citep{chen2016xgboost} is a scalable end-to-end tree boosting system, which is widely used in data science and achieves excellent results.
XGBoost proposes a novel sparsity-aware algorithm for sparse data and a weighted quantile sketch algorithm for approximate tree learning.
It uses the second-order approximation of the convex loss function that can optimize the objective quickly.
LightGBM \citep{ke2017lightgbm} proposes novel Gradient-based One-Side Sampling (GOSS)  and Exclusive Feature Bundling (EFB) methods.
With GOSS, LightGBM can obtain a quite accurate estimation of the information gain with much smaller data size.
The EFB method
bundles  mutually exclusive features using a greedy algorithm, solving the data sparsity problem.
Tree-based classifiers are usually more preferred in investment-related research, mainly because they can be interpreted.
Soft Decision Tree \citep{frosst2017distilling}
takes the knowledge acquired by neural nets and expresses the knowledge in a model that relies on hierarchical decisions, creating a more explicable model.
To our knowledge, there is no previous approach applying these methods in predicting the future success of startup companies.
We leverage this recent progress and compare the performance of these methods in predicting the future success of startup companies

\section{Our Approach}
\label{sec:our_approach}

We define the concept of 'success'  to include  raising new funding,  being acquired, or going for an IPO.
To make the prediction closer to the reality of VC investment,
we further restrict the concept of future success to be in a defined time window (18 months).

\subsection{Problem Statement}

We formulate the problem of evaluating startup companies as a binary classification problem.
For each company $i$, we synthesis a set of variables ${\bm{x_i}} \in \mathbb{R}^m$ to evaluate its future success, $m$ is the size of features.
The selection of the features is discussed in detail in Sec \ref{subsec:predicting_variables}.
The label $y_i$ is assigned according to whether the company will succeed in the defined time window:
\begin{equation}
\label{eq1}
y_{i}=
\begin{cases}
1,& \text{company i will succeed in the time window.} \\
0,& \text{company i will not succeed in the time window.} 
\end{cases}
\end{equation}
Our data set can be denoted as: $\mathcal{D} = \{({\bm{x_i}}, y_i)\}$, $i = 1,2, ... , n$, $n$ is the size of the dataset.
Given the labeled sample, the machine learning methods learn the conditional probability of $y$ given $\bm{x}$, i.e. $p(y|{\bm{x}})$.
Given a new data sample with no label, this model then outputs a prediction $\hat{y}$ that places the sample into the class of success or failure.
\begin{equation}
\label{eq:threshold}
\hat{y}_i =  f({\bm{x_i}}) = 
\begin{cases}
1,& p(y_i=1|{\bm{x_i}}) \geq th\\
0,& p(y_i=1|{\bm{x_i}}) < th
\end{cases}
\end{equation}
where $th$ is a threshold  manually selected. In our experiment, th is 0.5 for fair comparison.

We then try to construct portfolios according to the suggestion of these algorithms.
The aim of constructing a portfolio is to select a subset $\mathcal{S} = \{c_1, c_2, ... , c_k\}$ of companies with size $k$, each $c_i$ corresponds to a company, 
such that we maximize the expected number of success companies in the subset $\mathcal{S}$ .
\begin{equation}
\label{eq:portfolios}
\max_{|\mathcal{S}|=k} \bm{E}(\sum y_i |\{{\bm{x_i}}, i \in \mathcal{S}\})
\end{equation}
If we assume that the success of each company is independent, $\mathcal{S}$ is consisted of the companies with the $k$ largest $p(y_i=1|{\bm{x_i}})$.

\subsection{Data Preprocessing and Time Window}

Our data sample is extracted using the daily CSV export of Crunchbase on October 20, 2020.
The full dataset contains 1,166,402 organizations and 351,236 venture deals.
We filter out the companies that the date of establishment is missing.
The companies founded before 1990 are removed.
Unique ids are assigned to each company to distinguish from the duplication of company names.
Total 776,273 organizations remain after this preprocessing.

To make the prediction closer to the reality of VC investment,
we restrict the concept of future success to be in a defined time window.
We expect that the company will raise new funding, be acquired, or go for an IPO within a time threshold after the prediction.
These evaluation time windows can be interpreted as the time intervals for investors to evaluate the return of investments.
We use multiple time windows to 
enrich the number of data samples and take more factors like macroeconomy into consideration.

In practice, most VC firms believe the time segment between two funding rounds is usually around 18 months. 
Table \ref{table:time_window} summarizes the time startup companies need to raise next round funding,
validating that more than half of the companies achieve their next round within 18 months in most of the rounds.
So we define the evaluation time window of 18 months.
The time intervals and the according label distribution are shown in Table \ref{table1}
\footnote{There might be survival bias since the Crunchbase was founded in 2007. The companies failed before the creation of Crunchbase may not register in the dataset.}.
The time $t_s$ denotes the start of the evaluation window, which is the time we make the prediction and can be considered as the moment VC investors invest in a company.
The time $t_f$ denotes the end of the evaluation time window.
The companies that were acquired, went for an IPO, closed, or had no funding events before $t_s$ are removed.
Accumulated in all the time windows, the final data sample consists of 398,489 sample events.

\begin{table}[htbp]
\centering
\setlength{\belowcaptionskip}{3pt}%
\caption{Fundraising time interval. }
\label{table:time_window}
\begin{tabular}{p{70pt}<{\centering} p{60pt}<{\centering} p{60pt}<{\centering} p{60pt}<{\centering} p{60pt}<{\centering} }
\toprule
Funding round &  Mean interval (months) & Median interval (months) & 90th percentile (months)  & Within 18 months  \\
 \midrule
Seed $\rightarrow$ A&  22 & 18 &  43 & 52.28\%  \\
A $\rightarrow$ B&  22 & 18 & 40 & 51.15\% \\
B $\rightarrow$ C&  22 & 19 & 40 & 48.50\%  \\
C $\rightarrow$ D& 22 & 18 & 40 & 50.48\%  \\
D $\rightarrow$ E&  21 & 18 & 38 &  50.90\%  \\
E $\rightarrow$ F&  20 & 16 & 36 & 56.81\%  \\
F $\rightarrow$ G&  19 & 15 & 38 & 64.29\%  \\
G $\rightarrow$ H & 14 & 13 & 29 & 78.00\% \\
H $\rightarrow$ I & 11 & 10 &  35 & 90.00\%  \\
I $\rightarrow$ J&  10 & 8 & 19 & 66.67\%  \\
\bottomrule
\end{tabular}
\end{table}

\begin{table}[htbp]
\centering
\setlength{\belowcaptionskip}{3pt}%
\caption{The label distribution of different evaluation time windows}
\label{table1}
\begin{tabular}{p{70pt}<{\centering} p{70pt}<{\centering} p{50pt}<{\centering} p{50pt}<{\centering} p{50pt}}
\toprule
$t_s$&  $t_f$  &  Success&  Fail & Success\% \\
 \midrule
2000-01-01 & 2001-06-30 & 698 & 560 & 55.49\% \\
2001-07-01 & 2002-12-31 & 1008 &  3408 & 22.826\% \\
2003-01-01 & 2004-06-30  & 1385 & 4035 & 25.55\% \\
2004-07-01 & 2005-12-31 & 1734 & 4903 & 26.13\% \\
2006-01-01 & 2007-06-30 & 2442 & 5854 & 29.44\% \\
2007-07-01 & 2008-12-31 & 3118 & 8372 & 27.14\% \\
2009-01-01 & 2010-06-30 & 4017 & 11500 & 25.89\% \\
2010-07-01 & 2011-12-31 & 5667 & 15574 & 26.68\% \\
2012-01-01 & 2013-06-30 & 7340 & 21974 & 25.04\% \\
2013-07-01 & 2014-12-31 & 11370 & 30714 & 27.02\% \\
2015-01-01 & 2016-06-30 & 15833 & 45646 & 25.75\% \\
2016-07-01 & 2017-12-31 & 18851 & 66347 & 22.13\% \\
2018-01-01 & 2019-06-30 & 21046 & 85093 & 19.83\% \\
 \midrule
\multicolumn{2}{ c }{Total} & 94509 & 303980 & 23.72\% \\
 \midrule
\multicolumn{2}{ c }{Train} & 85064 & 273576 & 23.72\% \\
\multicolumn{2}{ c }{Train (SMOTE)} & 273576 & 273576 & 50.00\% \\
\multicolumn{2}{ c }{Test} & 9445 & 30404 & 23.70\% \\
\bottomrule
\end{tabular}

\end{table}

\subsection{Factor Exploration}
\label{subsec:predicting_variables}

Crunchbase provides information about companies, news, founders, funding rounds, and acquisitions.
We compile a set of 19 factors grouped in three categories related to the growth of companies based on the information available in the dataset, as summarized in Table \ref{table:factors}.
Since we use multiple time windows, all the factors are associated with the beginning of the evaluation time window ($t_s$).
Some variables that may have changed with time are omitted in our analysis.

\begin{table}[htbp]
\centering
\setlength{\belowcaptionskip}{3pt}%
\caption{The 19 Factors}
\label{table:factors}
\begin{tabular}{p{57pt} p{150pt} p{160pt}}
\toprule
Type & Factor & Definition \\
 \midrule
\multirow{7}*{\makecell[l]{Basic \\ information \\ \\ \\ \\ \\ }}  & com found year & number of years from 1990 to the year the company founded \\ 
                                                    & macroeconomy & number of newly established companies in the founding year \\ 
                                                    & company age & how long has the company been founded at $t_s$ (in month) \\ 
                                                    & number of news & total number of news before $t_{s}$ \\ 
                                                    & monthly average number of news &  monthly average number of news from the founding year to $t_{s}$ \\ 
                                                    & province(city) prosperity &  number of companies headquartered in the area at  $t_{s}$ \\  
                                                    & mean(max) province(city) prosperity of industries & average and max local prosperity of all industries associated with the company at $t_{s}$ \\
 \midrule
\multirow{4}*{\makecell[l]{Fundings \\ and \\ investors}}  & number of funding rounds & number of funding rounds the company achieved before $t_{s}$ \\ 
                                                              & total amount raised (in USD) & total amount raised in USD before $t_{s}$ \\ 
                                                              & mean(max) IPO fraction & average and max IPO fraction of all the investors of the company before $t_{s}$ \\
                                                              & mean(max) acquisition fraction & average and max acquisition fraction of all the investors of the company before $t_{s}$  \\
 \midrule
Founders & mean(max) founder fail fraction & average and max fail fraction of each founder before $t_{s}$ \\
\bottomrule
\end{tabular}
\end{table}

The first group of factors is about the general information of companies.
The year the company founded and the economic situation of that year are important for startup companies \citep{holmes2010analysis}.
Since we only consider companies founded after 1990 in our dataset, we use the years elapsed since 1990 to denote the founding year for convenience.
We take the number of newly established companies in the founding year as an indicator of the macroeconomy.
For different time windows, the company age at $t_s$ is an important indicator of the status of a company.
It is counted in months in our analysis.
News-related factors are useful measures of company performance \citep{xiang2012supervised}.
We count the total number of news and its total number of news  associated with each company.
The information of geographic environment  \citep{porter2001innovation,hoenen2012patents} and business sectors \citep{clarysse2011impact} are also very important for the development of startup companies.
We quantify the prosperity of an area by the number of companies headquartered in the area registered in Crunchbase.
Each company is associated with several industries in Crunchbase. 
The local prosperity of an industry is quantified by the number of companies associated with the industry in the area.
We calculate these factors with the geographical granularity of province and city.
When associating the prosperity of the industries to the company, we calculate the average and max local prosperity of all industries associated with the company.

The second group of factors is related to funding rounds and investors.
A startup company receives funding in a sequence of rounds.
The past funding experience is very important for both startup companies and venture capital firms  \citep{nahata2008venture, nanda2020persistent}.
We calculated the number of funding rounds the company achieved and the total amount raised in USD before $t_{s}$.
Research \citep{nahata2008venture} shows that reputable VC firms are more likely to lead their companies to successful exits.
The reputation of past investors is evaluated based on their historical investment data.
For each investor, we define the IPO fraction as the number of venture deals the investor invested in and exited with an IPO before $t_{s}$ divide by the total number of venture deals of the investor.
The definition of the acquisition fraction is similar to the IPO fraction, except we use the fraction of venture deals exited with an acquisition.
A company often has more than one investors, we utilize the average and max IPO and acquisition fraction of all its investors.

The third group of factors is about the founders.
The experience of the founders  \citep{jenkins2014individual, littunen2010rapid} and 
the founding team composition  \citep{nann2010power,eesley2014contingent}
 are also indicators that may influence the potential success of a company.
We define the fail fraction as the fraction that a founder had previously founded companies and failed before $t_{s}$.
As a company may have more than one founder, we consider the average and max fail fraction of all founders.

\subsection{Models and Algorithms }

We test eight different machine learning classifier algorithms: 
Logistic Regression, K Nearest Neighbor, Decision Tree, Multilayer Perceptron, 
Random Forests, XGBoost \citep{chen2016xgboost}, LightGBM \citep{ke2017lightgbm}, and soft Decision Tree \citep{frosst2017distilling}.

We use 90\% of the sample in the training phase, 10\% for the testing phase.
For the training dataset,
Synthetic Minority Over-sampling TEchnique (SMOTE) \citep{chawla2002smote} is used to handle the class imbalance problem, as shown in Table \ref{table1}.
SMOTE over-samples the minority class by taking each minority class sample and introducing synthetic examples along the line segments joining the k (k=5 in this paper) minority class nearest neighbors.
Except for over-sampling methods, some algorithms can handle the class imbalance problem by adjusting weights inversely proportional to class frequencies to the samples in the minority class.
For tree-based models (Decision Tree, Random Forests, XGBoost, and LightGBM ) the weights are adjusted in the calculation of split gain.
For Logistic Regression, Multilayer Perceptron, and soft Decision Tree, the weights are adjusted in the calculation of the loss function.
To optimize the hyperparameters with high  efficiency, we use Bayesian optimization to tune the hyperparameters.
We choose the best-performed hyperparameters for the experiments, some of the principal parameters are summarized in Table \ref{table:hyper_parameters}.

\begin{table}[htbp]
\setlength{\belowcaptionskip}{3pt}%
\caption{Hyperparameters}
\centering
\label{table:hyper_parameters}
\begin{tabular}{p{120pt} p{220pt} }
\toprule
Model&   Parameters \\
 \midrule
Random Forests& 133 estimators, max depth 63  \\
XGBoost&  180 estimators, max depth 11 \\
LightGBM&  355 estimators, max depth 8\\
soft Decision Tree&  tree depth 8, average weighted \\
Multilayer Perceptron & 2 hidden layers with 64 neurons in each, ReLU activation function, dropout rate 0.1 \\
\bottomrule
\end{tabular}
\end{table}

\section{Experiments and Results}
\label{sec:results}

\subsection{Performance Metrics}

The following parameters are used in computing the performance metrics:
\begin{itemize}
\item Recall (True Positive Rate, TPR): the percentage of correctly predicted successful companies to all successful companies in reality.
\item Precision: the percentage of correctly predicted successful companies to the total companies classified as successful by the classifier.
\item F1-measure: the weighted average of precision and recall.
\item False Positive Rate (FPR): the proportion of the failed companies got incorrectly classified by the classifier.
\end{itemize}

ROC curve (receiver operating characteristic curve) \citep{swets1988measuring}, 
a graph summarizing classifier performance of a classification model over a range of tradeoffs between TPR and FPR,
is used to compare the discriminative power of different models. 

\subsection{Experimental Results}

\begin{figure}[htbp]
\centering
\includegraphics[width=0.51\textwidth,]{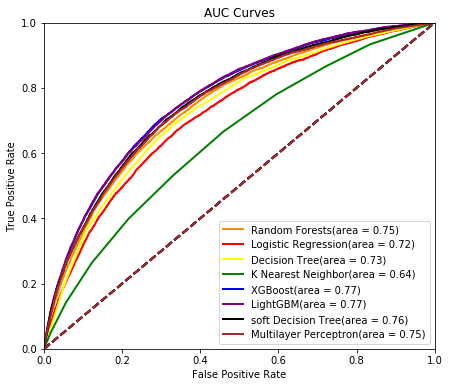}
\caption{ROC curves of different models.}
\label{fig:roc}
\end{figure}

The results of different machine learning classifier algorithms are shown in Table \ref{table:result_all}.
The results without SMOTE or weight adjustment are shown in Table \ref{table:result_all} (a).
Due to the class imbalance problem, the results tend to have relatively higher precision and lower recall.
The results with SMOTE over-sampling are shown in Table \ref{table:result_all} (b).
SMOTE can alleviate the class imbalance problem and improve the recall metric to some extent.
Adjusting weight balance in the model leads to better results in most of the models, as shown in Table \ref{table:result_all} (c).
The results show that LightGBM and XGBoost with weight adjustment performs best among eight machine learning methods,
achieving 53.03\% and 52.96\% F1, respectively.
The data of startup companies are usually incomplete for startup companies, especially for early-stage companies.
LightGBM and XGBoost are sparsity-aware algorithms and efficiently solve this data sparsity problem.
The ROC curves of different models are shown in Fig \ref{fig:roc}.

\begin{table}[htbp]
\caption{Results of predicting success }
\centering
\label{table:result_all}

\subtable[Baseline]{
\label{table:result2}
\begin{tabular}{ p{120pt}<{\centering} p{50pt}<{\centering} p{50pt}<{\centering} p{50pt}<{\centering} }
\toprule
Model&  Precision&  Recall&  F1 \\
\midrule
Logistic Regression& 0.6102  & 0.1372 & 0.2240 \\
K Nearest Neighbor& 0.4440 & 0.1532 & 0.2278  \\
Decision Tree& 0.5643 & 0.2242 & 0.3210 \\
Random Forests& \textbf{0.6558}   & 0.1297  & 0.2166  \\
XGBoost&  0.6050 & 0.2825 & 0.3851  \\
LightGBM& 0.6251 & 0.2680 &  0.3751 \\
soft Decision Tree&  0.6176 & 0.2274  & 0.3324 \\
Multilayer Perceptron &  0.5774 &  \textbf{0.2919} & \textbf{0.3878}  \\
Random Selection & 0.2372 & 0.5000 & 0.3218 \\
\bottomrule
\end{tabular}
}

\subtable[SMOTE]{
\label{table:result2}
\begin{tabular}{ p{120pt}<{\centering} p{50pt}<{\centering} p{50pt}<{\centering} p{50pt}<{\centering} }
\toprule
Model&  Precision&  Recall&  F1 \\
\midrule
Logistic Regression& 0.3730  &  0.6908 & 0.4844  \\
K Nearest Neighbor& 0.3303  & 0.5363 & 0.4088  \\
Decision Tree&   0.3976 & 0.5067  & 0.4456   \\
Random Forests&  0.4272  & 0.4940 & 0.4581 \\
XGBoost&  \textbf{0.5815} &  0.3185 & 0.4115 \\
LightGBM& 0.5770 & 0.3166  & 0.4088  \\
soft Decision Tree& 0.4149 & 0.6872 & \textbf{0.5174}  \\
Multilayer Perceptron & 0.4021 & \textbf{0.7254} & \textbf{0.5174} \\
Random Selection & 0.2372 & 0.5000 & 0.3218 \\
\bottomrule
\end{tabular}
}

\subtable[Weight adjustment]{
\label{table:result3}
\begin{tabular}{p{120pt}<{\centering} p{50pt}<{\centering} p{50pt}<{\centering} p{50pt}}
\toprule
Model&  Precision&  Recall&  F1 \\
\midrule
Logistic Regression&  0.3942 & 0.6813  & 0.4994 \\
K Nearest Neighbor&  --- &  --- &  --- \\
Decision Tree& 0.3928 & 0.6932  & 0.5014   \\
Random Forests&  0.4012  & \textbf{0.7154} & 0.5141 \\
XGBoost&  \textbf{0.4276} & 0.6955 & 0.5296\\
LightGBM& 0.4215 &  0.7150 & \textbf{0.5303}  \\
soft Decision Tree& 0.4145 & 0.6868 & 0.5170 \\
Multilayer Perceptron & 0.4129 & 0.6902 & 0.5167 \\
Random Selection & 0.2372 & 0.5000 & 0.3218 \\
\bottomrule
\end{tabular}
}
\end{table}

\subsection{Discussions on Single and Multiple Time Windows}

We define multiple time windows to enrich the number of data samples and to conduct time-aware analysis.
There may be a potential risk that the events that happen later in the training set can influence the prediction of earlier events, resulting in lower prediction power for future events.
To eliminate this concern, we conduct several experiments comparing models using single and multiple time windows to validate the robustness in predicting current and future events. 
 We use LightGBM for the experiments.

\begin{figure}[htbp]
\centering

\subfigure[In-sample period]{
\begin{minipage}[t]{0.5\linewidth}
\centering
\includegraphics[width=\textwidth]{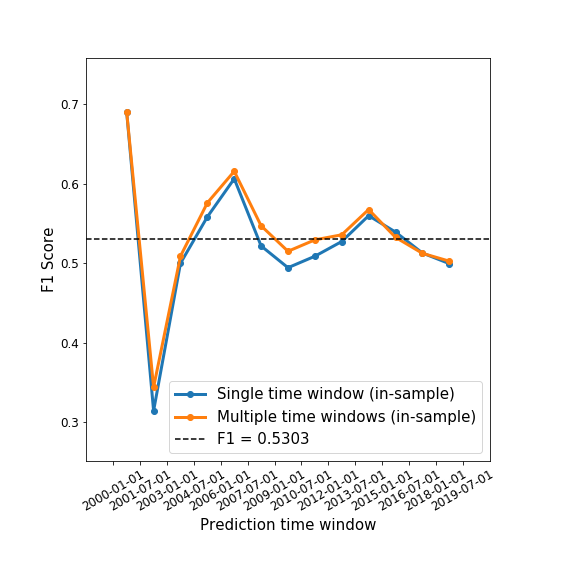}
\end{minipage}%
}%
\subfigure[Out-of-sample period]{
\begin{minipage}[t]{0.5\linewidth}
\centering
\includegraphics[width=\textwidth]{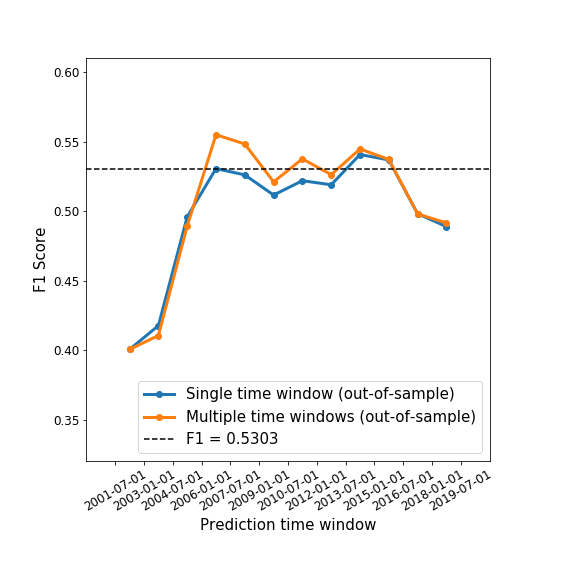}
\end{minipage}%
}

\caption{Comparing single and multiple time windows using LightGBM.}
\label{fig:single_time_window}
\end{figure}

First, we compare the performance of models using single and multiple time windows in predicting in-sample current events.
To do this, we use the test set in the last time window to evaluate the performance of the models.
Take the time window from January 2009 to June 2010 as an example.
For the single time window scenario, we take the 15517 sample events in this time window. Then we split 90\% of them as the training set and 10\% as the test set.
For the multiple time window scenario, we take 37517 sample events from all the time windows before January 2009 and combine them with the training set of the current time window to form the training set.
For a fair comparison, the test set of the multiple time window scenario is the same as the single time window scenario.
For each time window, we train two models using LightGBM with single and multiple time windows scenarios. 
We use Bayesian optimization to tune the hyperparameters of each model.
The result shows in Fig \ref{fig:single_time_window} (a).
Since the time windows start from January 2000, the results of the two scenarios in the first time window are the same.
The performance with multiple time windows scenario is slightly better than the single time window scenario, showing that adding historical data helps to improve the performance.
The difference is larger in earlier time windows when the amount of sample events is small.
After July 2013, there are more than 40,000 startup companies in a single time window, which is large enough to provide enough information and diversity for the prediction.
Thus the performance of the two scenarios is similar in recent time windows.

We also compare the performance of models using single and multiple time windows in predicting out-of-sample future events.
For this purpose, we use the models trained with historical time windows to predict the future success of the next time window.
For example, to predict the future success of the 15517 sample events in the time window from January 2009 to June 2010, we use the models trained on the datasets of time windows before January 2009.
For the single time window scenario, we take the 11490 sample events in the time window from July 2007 to December 2008 to form the training set.
For the multiple time window scenario, we take the 37517 sample events merging all the time windows before January 2009 to form the training set.
The result shows in Fig \ref{fig:single_time_window} (b).
Since the time windows start from January 2000 the results of these two scenarios in the first time window are the same.
The performance with multiple time windows scenario is slightly better than the single time window scenario.
This result agrees with the experiment of in-sample events, showing that adding historical data has a positive impact on the extrapolation to predict future events.

\subsection{Factor Importance }
\begin{figure}[htbp]
\includegraphics[width=0.8\textwidth,angle=-90]{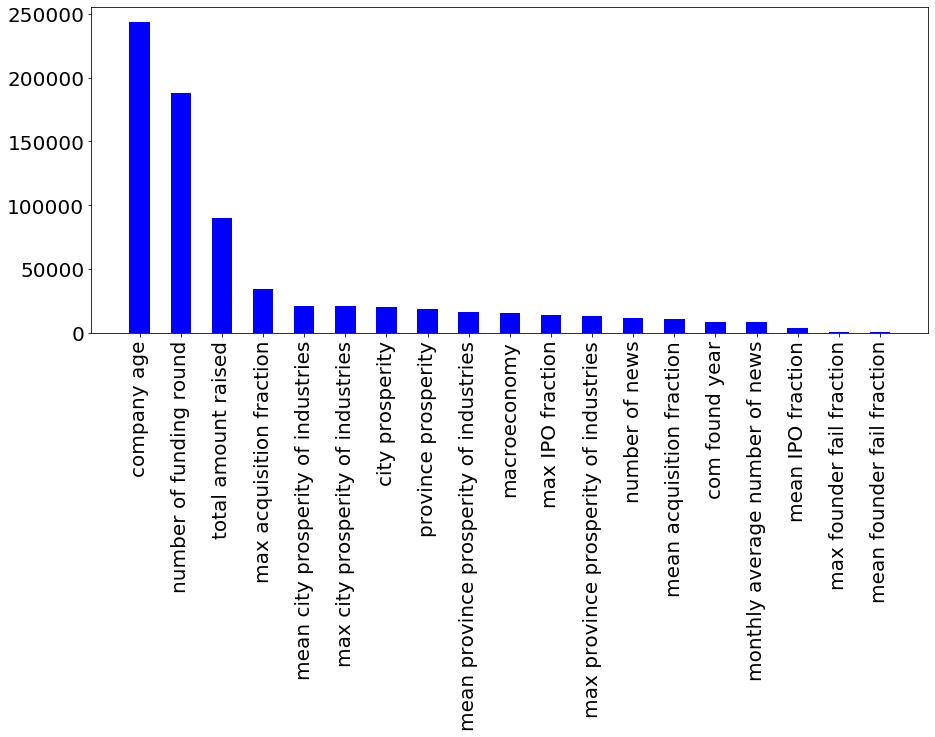}
\caption{Feature importance of LightGBM}
\label{fig:feature_importance}
\end{figure}

Based on the experimental results, we use LightGBM
to explore the importance of different factors.
The factor importance is calculated as the total gains of splits which use the feature.
The higher the value the more important and predictive the factor.
As shown in Fig \ref{fig:feature_importance},
the most important factors are company age and past funding experience.
The reputation of past investors, local prosperity, macroeconomy, and news also have some predictive power,
while the experience of the founders has little influence on the future success of the companies.

\subsection{Interpreting Model Predictions}
When an investor decides whether to take actions based on a prediction, understanding why a model makes a certain prediction is also important.
Explaining the reasons behind the prediction provides more insights into the model and makes the prediction more convincing.
SHAP (SHapley Additive exPlanations) \citep{NIPS2017_7062},
a game-theoretic approach to explain the output of any machine learning model,
is used to assign each feature an importance value for each prediction. 

\begin{figure}[htbp]
\centering
\includegraphics[width=\textwidth]{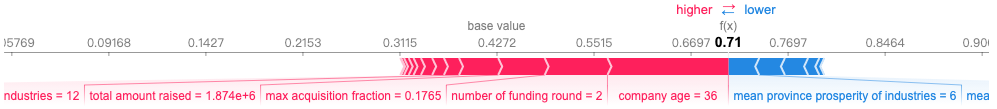}
\caption{Prediction interpretation of Market Logic Software. The graphic was generated by the Python SHAP library.}
\label{fig:shap_value}
\end{figure}

We take the company Market Logic Software as an example.
The output of LightGBM based on the data on 2008-12-31 is 0.71, indicating that the company had a high success probability in the following 18 months (2009-01-01 to 2010-06-30)\footnote{Actually, Market Logic Software raised its next round on May 5 2010.}.
Figure \ref{fig:shap_value} shows how each factor contribute to this output.
The base value is the average model output sampled from the training dataset.
Features pushing the prediction higher are shown in red, while those pushing the prediction lower are in blue.
As shown in the figure, company age, the number of funding rounds and reputable investors have positive effects on the future success of the company, 
local prosperity of its industrial sector is the main drawback.

\section{Portfolio Construction and Model Validation}
\label{sec:portfolio}

\begin{figure}[htbp]
\centering
\includegraphics[width=0.8\textwidth]{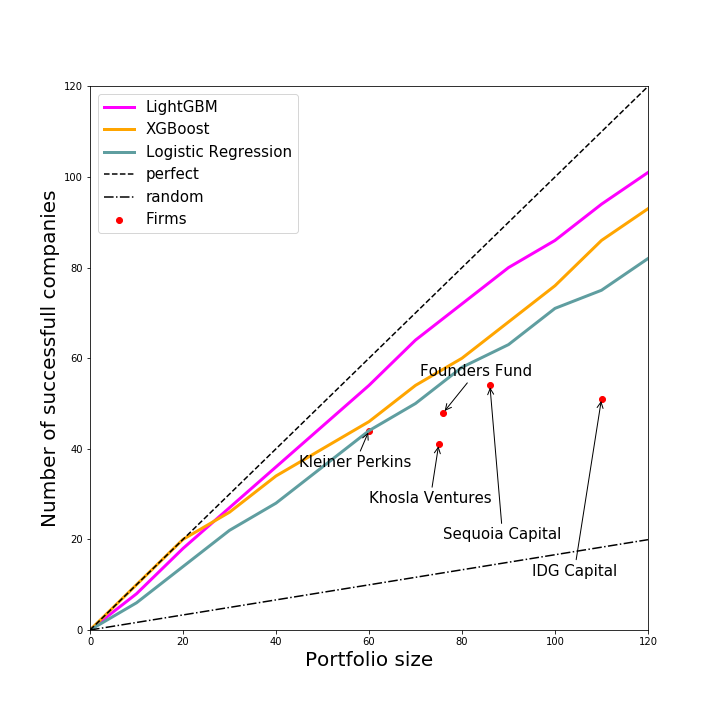}
\caption{Performance curve for portfolios constructed from different models.}
\label{fig:portfolio_exit}
\end{figure}

To validate that the models trained on current values can predict the future success of companies, we validate the models with out-of-sample periods.
We calculate the conditional probability of the success of each company in the time window from Jan 1st, 2019 to Jun 30, 2020,
using their features on Dec 31, 2018.
This out-of-sample set is composed of 121462 companies, with 20184 of them are successful.
This time window is later than the last time window in the training.
We construct portfolios of different sizes with LightGBM, XGBoost, and Logistic Regression
according to Eq. \ref{eq:portfolios}.
We use the number of successful companies in the portfolios to evaluate the predictive power of these machine learning models.
The number of companies that succeed in the portfolio versus the portfolio sizes is shown in Figure \ref{fig:portfolio_exit}.
The LightGBM and XGBoost model performs better comparing to Logistic Regression, which agrees with the results of the in-sample period test dataset,
showing that  these two algorithms have a great generalization and extrapolation power in predicting future events.
In the portfolio constructed by LightGBM of size 10, 8 succeed in the following 18 months, as listed in Table \ref{table:company_list_lgb}.

The performance of several top venture capital firms are also shown in Figure \ref{fig:portfolio_exit}.
The portfolio sizes of these points are the number of companies that the firm invests in 2018.
This result shows that the success rates of machine learning models are better comparing to human experts.

\begin{table}[htbp]
\setlength{\belowcaptionskip}{3pt}%
\caption{Top 10 companies selected by LightGBM}
\label{table:company_list_lgb}
\centering
\begin{tabular}{ p{70pt}<{\centering} p{100pt} p{60pt}<{\centering} p{100pt} }
\toprule
Company & Last Venture deal before 2019 &  Success probability& First venture deal  from 2019-01-01 to 2020-06-30   \\
\midrule
Lyft & Series I on Jun 28, 2018 & 0.9751 & IPO on Mar 29, 2019  \\
Coinbase & Series E on Oct 30, 2018 & 0.9667 & No Event  \\
Grab  & Series H on Dec 12, 2018 & 0.9593 & Series H  on Jan 7, 2019 \\
Revolut & Series C on  Apr 26, 2018 & 0.9590 & Non Equity Assistance on Mar 27, 2019   \\
Wealthsimple & Venture Round on Feb 22, 2018 & 0.9558 & Venture Round on May 22, 2019  \\
Improbable & Corporate Round on Jul 26, 2018  & 0.9549 & No Event \\
Privitar & Corporate Round on Dec 10, 2018 & 0.9546 & Series B on Jun 10, 2019  \\
Stripe & Series E on Sep 27, 2018 & 0.9529 & Series E+ on Jan 29, 2019  \\
Kabbage &  Debt Financing on Nov 16, 2017 & 0.9513 & Debt Financing on Apr 8, 2019  \\
BigBasket  & Series E on Jul 18, 2018 & 0.9507 & Series F on May 6, 2019 \\
\bottomrule
\end{tabular}
\end{table}

\begin{figure}[htbp]
\centering
\subfigure[Before Series A]{
\begin{minipage}[t]{0.5\linewidth}
\centering
\includegraphics[width=\textwidth]{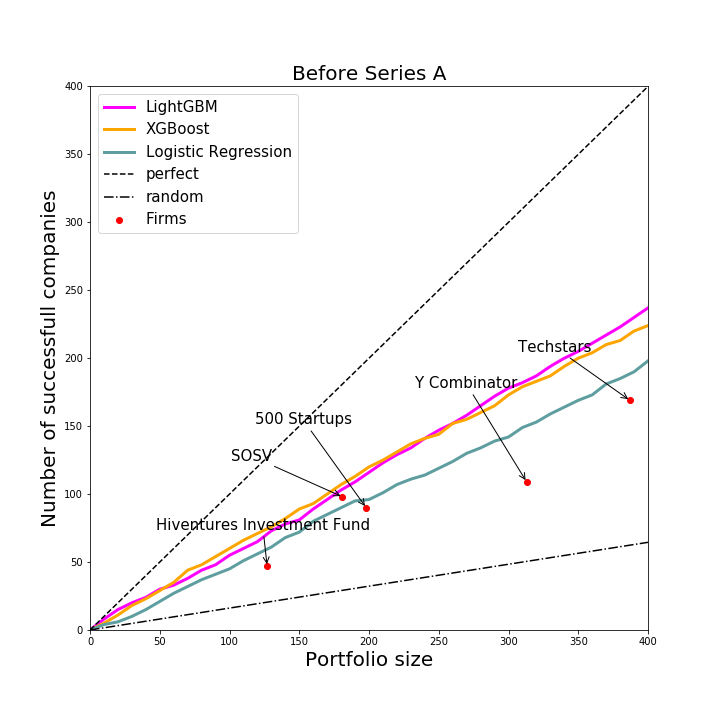}
\end{minipage}%
}%
\subfigure[Series A]{
\begin{minipage}[t]{0.5\linewidth}
\centering
\includegraphics[width=\textwidth]{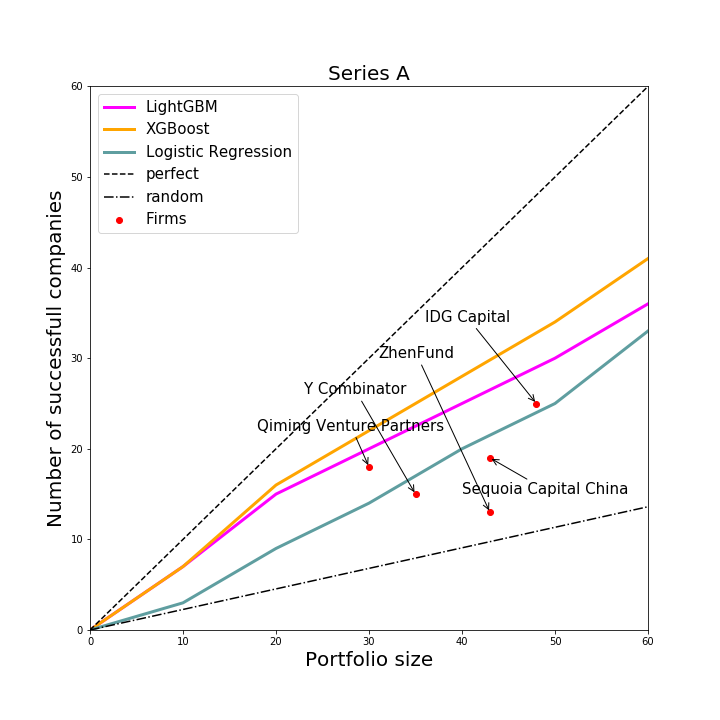}
\end{minipage}%
}
\subfigure[Series B]{
\begin{minipage}[t]{0.5\linewidth}
\centering
\includegraphics[width=\textwidth]{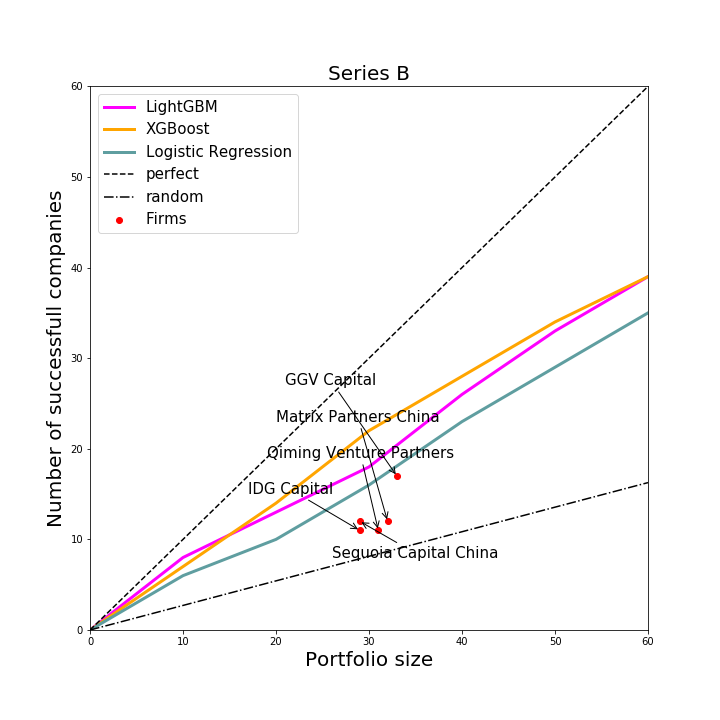}
\end{minipage}%
}
\caption{Performance curves for portfolios constructed in the early stages.}
\label{fig:portfolio_exit_round}
\end{figure}

Most of the companies listed in Table \ref{table:company_list_lgb} are in their late stages.
To further validate the robustness of our models in the early stages, we test the predictive power of the models in early-stage companies. 
We select companies in investment stages before Series A, Series A, and Series B according to their last funding round before 2019.
The top-ranking companies are listed in  Tables \ref{table:company_list_lgb_before_series_a}, \ref{table:company_list_lgb_series_a}, and \ref{table:company_list_lgb_series_b}.
The comparison with several top venture capital firms is shown in Figure \ref{fig:portfolio_exit_round}.
The performance of machine learning models is still comparable to human experts.

\begin{table}[htbp]
\setlength{\belowcaptionskip}{3pt}%
\caption{Top 10 companies selected by LightGBM (Before Series A)}
\label{table:company_list_lgb_before_series_a}
\centering
\begin{tabular}{p{70pt}<{\centering} p{100pt} p{60pt}<{\centering} p{100pt} }
\toprule
Company & Last Venture deal before 2019 &   Success probability& First venture deal  from 2019-01-01 to 2020-06-30  \\
\midrule
Electron & Seed Round on  Nov 20, 2018 &  0.9503 & Seed Round on Mar 1, 2019  \\
ACTO & Seed Round on Jan 18, 2018  &  0.9379 & No Event  \\
HqO & Seed Round on Sep 27, 2018 &  0.9322 & Series A on Feb 8, 2019  \\
Tomorrow Ideas & Convertible Note on Oct 3, 2018  &  0.9315 & Venture Round on  Nov 1, 2019 \\
Optimal & Seed Round on Dec 22, 2017   & 0.9258 & No Event \\
FunnelAI & Non Equity Assistance on Feb 21, 2018  &  0.9224 & Seed Round on Mar 27, 2019 \\
Sprout.ai & Pre Seed Round on Apr 2, 2018  &  0.9213 & Seed Round on Jun 1, 2019 \\
ICON & Seed Round on Oct 17, 2018  &  0.9195 &  Venture Round on Jan 23, 2020   \\
Lifebit & Seed Round on Jul 19, 2018  &  0.919355 & Seed Round on Apr 30, 2020 \\
Liveoak Technologies & Seed Round on Sep 13, 2017  & 0.9183 & Series A on Jun 4, 2019    \\
\bottomrule
\end{tabular}
\end{table}

\begin{table}[htbp]
\setlength{\belowcaptionskip}{3pt}%
\caption{Top 10 companies selected by LightGBM (Series A)}
\label{table:company_list_lgb_series_a}
\centering
\begin{tabular}{p{100pt}<{\centering} p{100pt}  p{60pt}<{\centering} p{100pt} }
\toprule
Company & Last Venture deal before 2019 &  Success probability& First venture deal  from 2019-01-01 to 2020-06-30  \\
\midrule
Honeycomb & Series A on Feb 1, 2018 & 0.9357 & Series A+ on Sep 26, 2019  \\
League Network $\vert$ MyDrCares  $\vert$ FundMyTeam  & Series A on Jan 12, 2018 & 0.9297 & No Event  \\
Bark Technologies & Series A on Aug 29, 2018 & 0.9259 & Series B on Mar 1, 2020 \\
Hometree & Series A on  Sep 26, 2018 & 0.9241 & Venture Round on Jan 1, 2019  \\ 
Pypestream &  Series A+ on Dec 13, 2018  & 0.9202 & No Event \\
Inspectorio  & Series A on Jul 11, 2018 & 0.9134 & No Event \\
Triple W Japan & Series A+ on Nov 6, 2017 & 0.9111 & Venture Round on Jan 1, 2019  \\
Terminal & Series A on May 22, 2018  & 0.9104 & Series B on Sep 26, 2019 \\
Owkin & Series A+ on May 23, 2018  & 0.9100 & Series A++ on Mar 7, 2019 \\
Shipwell &   Series A on Oct 9, 2018 & 0.9098 &  Series B on Oct 24, 2019  \\
\bottomrule
\end{tabular}
\end{table}

\begin{table}[htbp]
\setlength{\belowcaptionskip}{3pt}%
\caption{Top 10 companies selected by LightGBM (Series B)}
\label{table:company_list_lgb_series_b}
\centering
\begin{tabular}{p{70pt}<{\centering} p{100pt}  p{60pt}<{\centering} p{100pt} }
\toprule
Company & Last Venture deal before 2019  &  Success probability& First venture deal  from 2019-01-01 to 2020-06-30 \\
\midrule
BigID  & Series B on Jun 25, 2018 & 0.9307 & Series C on Sep 5, 2019 \\
Valimail, Inc. &  Series B on May 22, 2018 & 0.9274 & Series C on Jun 19, 2019  \\
Imperfect Foods & Series B on Jun 27, 2018 & 0.9231 & Series B+ on Mar 31, 2020 \\
Kayrros & Series B on Sep 18, 2018  & 0.9191 & No Event \\ 
Aircall  & Series B on May 15, 2018  & 0.9187 & Series C on May 27, 2020\\
Zero Hash  & Series B+  on Sep 12, 2018 & 0.9099 & No Event  \\
Thread & Series B on Oct 16, 2018  & 0.9074 & Series B+  on Nov 20, 2019 \\
Packet & Series B on Sep 10, 2018  & 0.9013 & Acquired by Equinix on Jan 15, 2020 \\
AXIOS Media & Series B on Nov 17, 2017  & 0.9010  & Series C on Dec 29, 2019  \\
Pagaya Investments &  Series B on Aug 30, 2018 & 0.8979 & Series C on Apr 3, 2019   \\
\bottomrule
\end{tabular}
\end{table}

Our models could help investors to decrease the failure rates in their portfolios.
The success rates of machine learning methods are high since we do not consider many practical factors, such as funding size, investment preference, whether the company is within reach, and so on.

\section{Conclusion and Discussion}
\label{sec:conclusion}

In this work, we try to solve the data sparsity problem with recent machine learning methods.
We analyze several machine learning methods
using a large dataset derived from CrunchBase.
We conduct a time-aware analysis based on multiple time windows, which is more practical in real-world scenarios.
We expand the scope of success that includes raising new funding,  being acquired, or going for an IPO.
The results show that the two sparsity-aware algorithms,
 LightGBM and XGBoost, 
 perform best among eight machine learning methods and achieve 53.03\% and 52.96\% F1 scores in the prediction, respectively.
Through feature mining,
we find that company age and past funding experience are among the most important factors.
We also interpret the predictions from the perspective of feature contribution.
We construct portfolio suggestions according to these methods with out-of-sample periods, which achieve better results compared to human experts.
The results show that our methods have great generalization and extrapolation power in predicting future events.
These findings have substantial implications on how machine learning methods can help investors identify potential business opportunities.

Future studies will include integrating more data sources and discovering more features, such as features related to founders, public opinions, and so on.
Instead of building a model to predict all companies, we will try to build multiple models according to sectors so that we can create customized features to help improve the performance of the prediction in different sectors.
We will also focus on the generalization and interpretation of the models.
such as introducing causal inference methods to extract the potential reasons for a successful prediction. 

\section*{Acknowledgement}
\label{sec:acknowledgement}
We would like to thank Jiren Zhu and Haomin Wang for their supports and great insights on success prediction. 
We would also like to thank Yaoqiang Xing for his supports on data management. 
We would like to thank Dr. Kaifu Lee for reviewing the paper and giving very illuminative suggestion. 
Our work would not have been possible without their support.

\vfill
\pagebreak



%%       \citet{<label>} ==> Jones et al. [21]
%%       \citep{<label>} ==> [21]
%%



%\begin{thebibliography}{00}


\bibliographystyle{elsarticle-harv.bst}\biboptions{authoryear}
\bibliography{mybibfile}

%\end{thebibliography}
\end{document}